\documentclass[lettersize, journal]{IEEEtran}
\usepackage{amsmath,amsfonts}
\usepackage{algpseudocode}
\usepackage{algorithm}
\usepackage{array}
\usepackage[caption=false,font=normalsize,labelfont=sf,textfont=sf]{subfig}
\usepackage{textcomp}
\usepackage[T1]{fontenc}
\usepackage{stfloats}
\usepackage{url}
\usepackage{verbatim}
\usepackage{graphicx}
\def\BibTeX{{\rm B\kern-.05em{\sc i\kern-.025em b}\kern-.08em
		T\kern-.1667em\lower.7ex\hbox{E}\kern-.125emX}}
\usepackage{balance}
\begin{document}
\title{Fact-based Agent modeling for Multi-Agent Reinforcement Learning}
\author{Baofu Fang, Caiming Zheng and Hao Wang
	\thanks{Manuscript received 4, June 2023. This work was supported by the University Synergy Innovation Program of Anhui Province (Grant No.GXXT-2022-055), Open Fund of Key Laboratory of Flight Techniques and Flight Safety, CAAC (Grant No.FZ2022KF09), and the R\&D Program of Key Laboratory of Flight Techniques and Flight Safety, CAAC(Grant No.FZ2022ZZ02).
	
	The authors are with the School of Computer Science and Information Engineering, Hefei University of Technology, Hefei, 230601, China (e-mail: fangbf@hfut.edu.cn; 2502282770@@qq.com; jsjwangh@hfut.edu.cn).
}}

\markboth{Journal of \LaTeX\ Class Files,~Vol.~18, No.~9, September~2020}%
{How to Use the IEEEtran \LaTeX \ Templates}
\maketitle

\begin{abstract}
	In multi-agent systems, agents need to interact and collaborate with other agents in environments. Agent modeling is crucial to facilitate agent interactions and make adaptive cooperation strategies. However, it is challenging for agents to model the beliefs, behaviors, and intentions of other agents in non-stationary environment where all agent policies are learned simultaneously. In addition, the existing methods realize agent modeling through behavior cloning which assume that the local information of other agents can be accessed during execution or training. However, this assumption is infeasible in unknown scenarios characterized by unknown agents, such as competition teams, unreliable communication and federated learning due to privacy concerns. To eliminate this assumption and achieve agent modeling in unknown scenarios, Fact-based Agent modeling (FAM) method is proposed in which fact-based belief inference (FBI) network models other agents in partially observable environment only based on its local information. The reward and observation obtained by agents after taking actions are called facts, and FAM uses facts as reconstruction target to learn the policy representation of other agents through a variational autoencoder. 
	We evaluate FAM on various Multiagent Particle Environment (MPE) and compare the results with several state-of-the-art MARL algorithms. Experimental results show that compared with baseline methods, FAM can effectively improve the efficiency of agent policy learning by making adaptive cooperation strategies in multi-agent reinforcement learning tasks, while achieving higher returns in complex competitive-cooperative mixed scenarios.
\end{abstract}

\section{Introduction}
\IEEEPARstart{R}{einforcement}
Learning (RL) has achieved rapid progress in cooperative and competitive multi-agent games, such as OpenAI Five\cite{berner2019dota} and AlphaStar\cite{vinyals2019grandmaster}. In multi-agent environments, agents must interact with each other, where the interaction relationship includes competition and cooperation. Due to the policy of all agents are simultaneously learning, it affects the state transitions and reward functions experienced by an individual agent\cite{yu2022model}. From the perspective of a single agent, interacting with other agents whose policies change makes the environment non-stationary. Therefore, other agents cannot be simply treated as part of the environment. Agent modeling promotes the agent to adjust its own policy to adapt to the policy changes of other agents by explicitly modeling the beliefs, behaviors and intentions of other agents\cite{albrecht2018autonomous}. Since the agent learns in the same partially observable environment while other agents whose strategies are complex, diverse, and time-varying. Therefore, modeling other agents in non-stationary environments is a major challenge for multi-agent reinforcement learning.

Traditional agent modeling assumes that agents can access the local information of other agents during execution and training\cite{albrecht2018autonomous,he2016opponent,hong2018deep}, including the local observations and actions taken by other agents. However, this assumption often does not hold in many multi-agent scenarios. In practical, agents may have limited visibility of their surrounding environment and communication with competing agents may be prohibited, while communication between cooperating agents is often unreliable\cite{chen2022local}, for example in federated learning tasks. In such situations, agents must inference and make decisions based on their local information. To weaken this assumption, LIAM\cite{Georgios2021Agent} and SMA2C\cite{Georgios2020Variational} utilize the local information of agents, including their own observations, actions, and rewards, to infer the representations of other agents in a recurrent manner. 
These methods relax the assumption of traditional agent modeling by allowing access to the local information of other agents only during the training stage. However, in unknown scenarios, agents may also be prohibited from accessing the information of other agents during both execution and training stages. It is infeasible for behavior cloning-based approaches to explicitly minimize the difference between an agent's policy model and the true policy. Therefore, the agent modeling in this case requires the agent to rely on its own local information, that is, it does not access the local information of other agents during the training and execution phases.

\begin{figure*}[!t]
	\centering
	\includegraphics[width=1\linewidth]{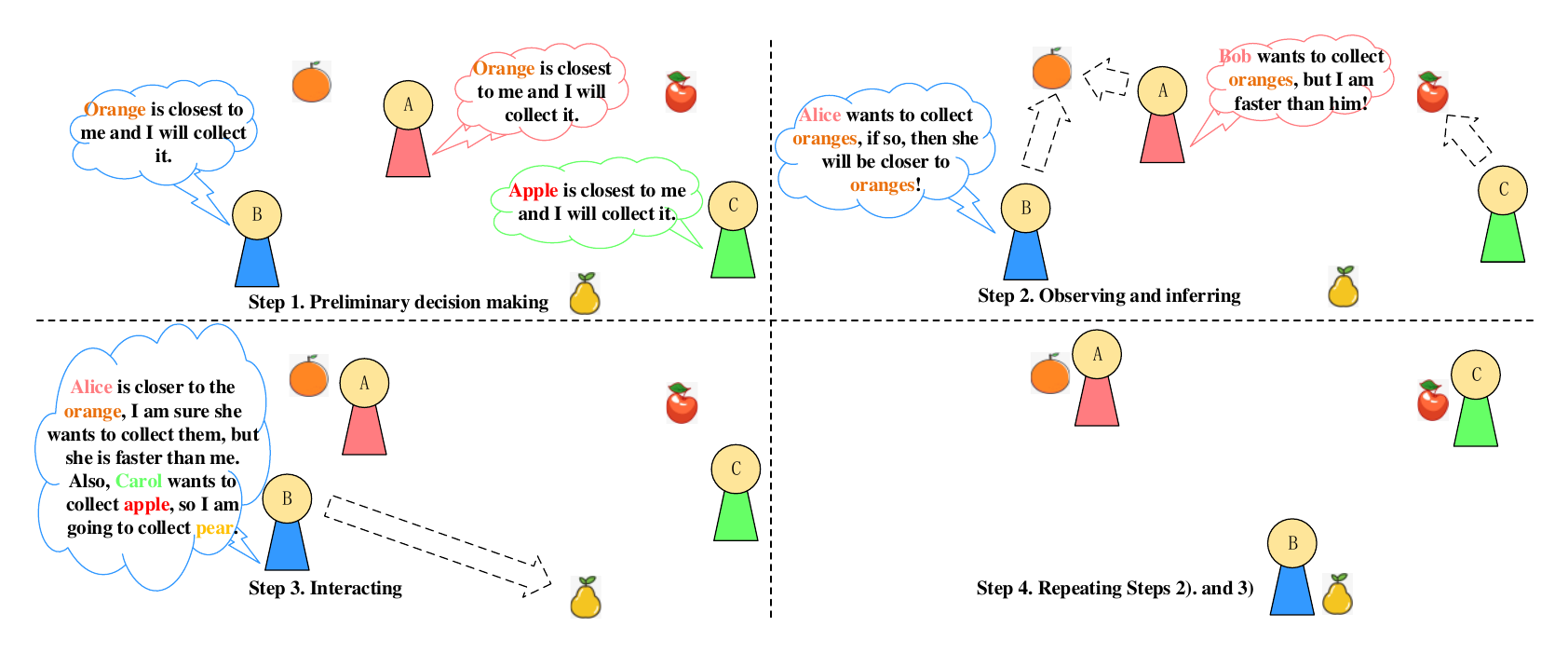}
	\caption{Fruit collection example. Everyone needs to cooperate to collect three konds of fruits which are apple, orange, and pear in shortest possible time. The whole process of fruit collection can be divided into 4 stages: 1).Preliminary decision making: Observing the surrounding environment, independently select fruits, and making effective decisions. 2).Observing and inferring: Observing information related others and inferring their behavioral intentions. 3).Interacting: Making adaptive decisions based on the inferred results to interact with environment and other agents and verifying the previous inferences through the fact of consequences after decisions. 4).Repeating steps 2). and 3). to avoid cooperative goal conflicts and achieve collaborative consensus until the fruit collection task is completed.}
	\label{fig:fruit_collection_example}
\end{figure*}

Consider a simple real-world scenario as shown in Figure \ref{fig:fruit_collection_example}, where the fruits collection task requires three people (Alice, Bob and Carol) to collect three fruits include apples, oranges, and pears with the shortest time. In order to achieve this task without communication, each person should go through four stages: 1).Preliminary decision making, 2).Observing and inferring, 3).Interacting, 4).Repeating steps 2) and 3) to avoid conflits and achieve collaborative consensus until the fruits collection task is complete. In this process, each person needs to start from the recent observation to infer other person policy representations to help itself make adaptive decisions. At the same time, the facts that happened after decision making are used to verify the inference result. In multi-agent systems, the rewards and observation received by the agent after performing the action imply rich information about the actions of other agents at the same moment. 

Based on this viewpoint, We propose Fact-based Agent Modeling (FAM) for multi-agent learning, which eliminates the assumption of accessing local information of other agents during execution and training phases. We build fact-based belief inference (FBI) network to model other agents based on own local information which is a variational autoEncoder (VAE) that has the advantage of being able to compensate for the information difference between the execution and training phases. The difference from the existing work is that SMA2C\cite{Georgios2020Variational} adopts the method of behavior cloning during training phase. Howerver, in this paper, the reward signal containing global information and the observation of local information received by the agent after performing the action are used as the reconstruction target. 
The proposed FAM is suitable for non-stationary and partially observable environments. In addition, the complexity of agent modeling of SMA2C\cite{Georgios2020Variational} and LIAM\cite{Georgios2021Agent} is $\mathcal{O}(N)$ while FAM is $\mathcal{O}(1)$ that is independent of the number of agents. FAM is also more suitable for unknown scearios. The main contributions of this article as follows. 

\begin{enumerate}
	\item{In order to elimate the assumption of accessing the local information of other agents for agent modeling, fact-based belief inference (FBI) network is proposed, which models other agents based on own local information using variational autoencoder.}
	
	\item{Combining FBI with Actor-Critc, Fact-based Agent Modeling (FAM) is proposed for multi-agent learning, which learns adaptive collaboration strategies by considering the policies of other agents. It can effectively applicable to partially observable environments.}
	
	\item{Extensive experimental was conducted to verify the effectiveness and feasibility of the proposed FAM, and analyze the information encoded by FBI.}
\end{enumerate}

The remainder of this article is organised as follows. Section II describes the background of deep reinforcement learning and variational autoencoder. Section III reviews the related work about multi-agent reinforcement learning and agent modeling. Section IV presents the proposed FBI and FAM. Section V describes the detailed results and analysis of the experiments. Ultimately, conclusions are provided in Section VI.

\section{Notation and Background}
The fully cooperative multi-agent task can be modeled as a decentralized partially observable Markov decision process(Dec-POMDP)\cite{oliehoek2016concise}. It is represented by the tuple $G = <\mathcal{I}, \mathcal{S}, \mathcal{U}, P, r, Z, \mathcal{O}, \gamma>$, where $\mathcal{I}=\{1,2,...,n\}$ is a finite set of agents, and $n$ represents the number of agents. $s\in \mathcal{S}$ describes the global state of the environment. At each timestep $t$, each agent $a\in \mathcal{I}$ receives an observation $o_{t}^{a}\in \mathcal{O}$ through the observation function $Z(s,a):\mathcal{S}\times \mathcal{I} \rightarrow \mathcal{O}$ and selects an action $u_{t}^{a}$, forming a joint action $\textbf{u}_{t} \in \mathcal{U}$. After executing the actions, the agents receive rewards signal $r_{t}$, where all agents share the same reward function $r(s,\textbf{u}):\mathcal{S} \times\mathcal{U}\rightarrow \mathbb{R}$, and transition to the next state according to transition probability function $P(s'|s,\textbf{u}):\mathcal{S} \times \mathcal{U}\times \mathcal{S}\rightarrow [0,1]$. The action-observation history of each agent is denoted as $\tau^{a}\in \mathcal{T}\equiv (\mathcal{O} \times \mathcal{U})^{*}$, and the policy $\pi(u_{t}^{a}|\tau_{1:t}^{a};\theta_{a}): \mathcal{T} \times \mathcal{U} \rightarrow [0,1]$ is based on its own action-observation history, with policy parameters $\theta_{a}$. The goal of Dec-POMDP is to learn a joint policy $\pi=(\pi_{1},...,\pi_{n})$ that maximizes the team cumulative discounted return $R_{t}=\sum_{l=0}^{\infty}\gamma^{l}r_{t+l}$, where $\gamma \in [0,1)$ is the discount factor. The joint action-value function of the joint policy $\pi$ is denoted as $Q^{\pi}(s_{t},\textbf{u}{t})=\mathbb{E}_{s_{t+1:\infty},\textbf{u}_{t+1:\infty}}\left[ R_{t} | s_{t},\textbf{u}_{t} \right]$, the joint state-value function is denoted as $V^{\pi}(s{t})=\mathbb{E}{s_{t+1:\infty},\textbf{u}_{t:\infty}}\left[ R_{t} | s_{t} \right]$, and the advantage function is denoted as $A^{\pi}(s_{t},\textbf{u}_{t})=Q^{\pi}(s_{t},\textbf{u}_{t})-V^{\pi}(s_{t})$.

\noindent \textbf{Policy Gradient}: Vanilla Policy Gradient (REINFORCE) is an on-policy algorithm that directly uses a parameterized model\cite{sutton1999policy} to approximate the policy $\pi(u_{t}|s_{t};\theta)$. REINFORCE does not require a separate behavior policy because $\pi_{\theta}(u_{t}|s_{t})$ naturally explores and exploits the environment. The policy parameters $\theta$ are updated at each step by increasing the log-likelihood of the chosen actions with respect to the sampled trajectory return $R_{t}$. The gradient update direction is given by:
\begin{equation}
	\label{eq:pg_actor_loss1}
	g = \mathbb{E}_{t} \left[
	\nabla_{\theta} \log \pi(u_{t}|s_{t};\theta) R_{t}
	\right]
\end{equation}

The baseline $b_{t}(s_{t})$\cite{williams1992simple} is subtracted from the return to reduce the variance of the estimated return while remaining unbiased, and the gradient update direction becomes:
\begin{equation}
	\label{eq:pg_actor_loss2}
	g = \mathbb{E}_{t}\left[
	\nabla_{\theta} \log \pi(u_{t} | s_{t};\theta)(R_{t}-b_{t}(s_{t}))
	\right]
\end{equation}

This allows for more stable learning and potentially faster convergence. The baseline can be a value function estimate or a learned function that approximates the expected return at state $s_{t}$.

\noindent \textbf{Advantage Actor-Critic (A2C)}: A2C is an on-policy Actor-Critic method that utilizes parallel environments to break the correlation between consecutive samples. It introduces a state value function estimator $V_{w}(s_{t})$ to approximate the state value $\mathbb{E}\left[ R_{t}|s_{t} \right]$, and used as a baseline to reduce the variance of sampling returns to improve policy gradient updates. Since $Q(s_{t},u_{t})$ is an approximate estimate of $R_{t}$, $A(s_{t},u_{t}) = R_{t}-V(s_{ t})$ is expressed as the advantage of action $a_{t}$ under state $s_{t}$, then the direction of A2C policy gradient update is as follows,
\begin{equation}
	\label{eq:a2c_actor_loss}
	g = \mathbb{E}_{t} \left[
	\nabla_{\theta} \log \pi(u_{t} | s_{t};\theta)(A(s_{t},u_{t}))
	\right]
\end{equation}
By using the advantage function $A(s_{t}, u_{t})$, A2C facilitates more stable and efficient learning. And the loss function for the state value function is given by:
\begin{equation}
	\label{eq:a2c_critic_loss}
	\mathcal{L}_{a2c}(\omega) = \mathbb{E}_{(s_{t},u_{t},r_{t+1},s_{t+1}) \sim B} (R_{t}-V_{\omega}(s_{t}))^{2}
\end{equation}
where $B$ denotes the sampled batch trajectory.

\noindent \textbf{Proximal Policy Optimization (PPO)}: PPO is an Actor-Critic algorithm whose core idea is to achieve stable training by limiting the distance between old and new policies. The PPO algorithm uses a loss function called "clipped surrogate objective" to control the step size of policy update, thus achieving stability in training without slowing down the training speed. Unlike A2C, PPO employs a technique called importance sampling, which allows multiple gradient descent updates to be performed using the same batch of trajectories. The loss function of actor for PPO as follows:
\begin{equation}
	\label{eq:ppo_critic_poss}
\mathcal{L}_{ppo}(\theta) = \mathbb{E}_{\tau \sim B} \left[ \min(\text{r}_{t}(\theta), \text{clip}(\text{ratio}_{t}(\theta),
1-\epsilon, 1+\epsilon)) A_{t} \right]
\end{equation}
where $\text{ratio}_{t}(\theta) = \frac{\pi_{\theta}(u_{t} | s_{t})}{\pi_{\theta_{old}}(u_{t} | s_{t})}$ represents the ratio between the new and old policies, and $\epsilon$ is a hyperparameter used to control the difference between the new and old policies. Compared to A2C, PPO has a higher sample efficiency.

\noindent \textbf{Variational Autoencoder (VAE)}: VAE is a generative model used to approximate the true posterior distribution $p(z|x)$, where the dataset samples $X=\{x_{i}\}_{i=1}^{N}$ are generated from an unknown parameterized generative distribution $p(x|z;\theta)$ with the latent variable $z$ being unobserved. The prior distribution of the latent variable $z$ is assumed to be a Gaussian distribution $p(z)=\mathcal{N}(z;0,I)$ with mean $0$ and variance $1$. The goal of VAE is to learn a variational distribution $q(z|x;\phi)$ parameterized by $\phi$ to approximate the true posterior distribution $p(z|x)$, where $q(z|x;\phi)=\mathcal{N}(z;\mu,\sigma,\phi)$ is a Gaussian distribution with mean $\mu$ and variance $\sigma$.

Variational inference uses the KL divergence as a distance measure function to minimize the distance between the approximate posterior distribution $q(z|x;\phi)$ and the true posterior distribution $p(z|x)$ , the Evidence Lower Bound(ELBO) as follows:
\begin{equation}
\begin{aligned}
\log p(x) &\ge ELBO(\psi,\phi|x) \\
&= \mathbb{E}_{z\sim q(z|x;\phi)} \left[
\log p(x|z;\psi)
- D_{KL}(q(z|x;\phi) || p(z))
\right]
\end{aligned}
\end{equation}
where $D_{KL}$ represents the Kullback-Leibler (KL) divergence. The first term on the right-hand side of the equation is the reconstruction loss, which measures the quality of the generated samples. The second term is a regularization term, which is used to constrain the distribution of the latent variables. Higgins et al.\cite{higgins2017beta} proposes $\beta$-VAE, where the parameter $\beta \ge 0$ is used to balance the reconstruction loss and the regularization term. The overall optimization objective of the $\beta$-VAE is as follows:
\begin{equation}
	L_{vae}(\phi,\psi) = \mathbb{E}_{z\sim q_{\phi}} \left[
	\log p(x|z;\psi) - \beta D_{KL}(q(z|x;\phi) || p(z))
	\right]
\end{equation}


\section{Related Work}
Multi-agent system (MAS) consists of multiple agents interacting in the shared environment to accomplish tasks. For complex tasks, Multi-Agent Reinforcement Learning (MARL) enables agents to learn effective policies through interaction with the environment. One of the main challenges in MARL is the inherent non-stationarity of the environment where all agents learn simultaneousl. Since the policies of other agents are unknown, it is unstable for agents to learn policies if they are considered part of the environment. To address this challenge, one approach is to adopt the Centralized Training with Decentralized Execution (CTDE) framework, where a centralized Critic is used to approximate joint action value or state-action value to guide the policy learning of individual agents. The value-based methods include QMIX\cite{rashid2020monotonic}, OW-QIX\cite{rashid2020weighted}, and TransfQMIX\cite{gallici2023transfqmix} and the Actor-Critic-based methods such as MADDPG\cite{lowe2017multi} and MAAC\cite{iqbal2019actor}.

Another approach to address the challenge of non-stationarity in MARL is agent modeling, which mitigates the effect of non-stationarity by incorporating information about other agents' beliefs, behaviors, and intentions. Many studies on agent modeling rely on predicting the actions or goals of other agents during training. He et al.\cite{he2016opponent} proposed a behavior cloning-based agent modeling approach that uses a neural network to predict the actions executed by other agents based on their observations. Hernandez-Leal et al.\cite{hernandez2019agent} treated learning other non-learning agents' policy as an auxiliary task and simplified it to a standard single-agent reinforcement learning problem. Similarly, Georgios et al.\cite{Georgios2021Agent} used an encoder to construct representations of other agents based on their local information in a recurrent manner, while the decoder reconstructed the observations and actions of other agents. Another method\cite{Georgios2020Variational} applied a variational autoencoder for agent modeling, where the encoder generates a high-dimensional continuous distribution as a representation of the other agent's policy, and the decoder is trained by reconstructing the agent's actions. Both of these methods allow access to other agents' local information during the training or execution.

In terms of considering simultaneously learning opponents, Foerster et al.\cite{foerster2018learning} proposed LOLA, which incorporates the influence of an agent's policy on the parameter updates of other agents' policies. Al-Shedivat et al.\cite{al2018continuous} introduced Mate-PG, a meta-policy gradient-based method that leverages the trajectories of other agents in multiple meta-gradient steps to construct a policy that benefits from updating other agents. Kim et al.\cite{kim2021policy} proposed an extension to the existing method called Meta-MAPG. They introduced an additional term that captures the influence of an agent's current policy on the future policies of other agents, similar to LOLA. These meta-learning-based methods require the trajectory distributions to match between training and testing, implicitly assuming that all other agents use the same learning algorithm.

Unlike existing work, we consider a more complex and general setting where the policies of other agents are learned simultaneously with the agent's own policy. Furthermore, there is partial observability in multi-agent environment, and the agents are not allowed to access the local information of other agents to achieve agent modeling during the execution and training.

\section{Methods}
\begin{figure}[!t]
	\centering
	\includegraphics[width=0.9\linewidth]{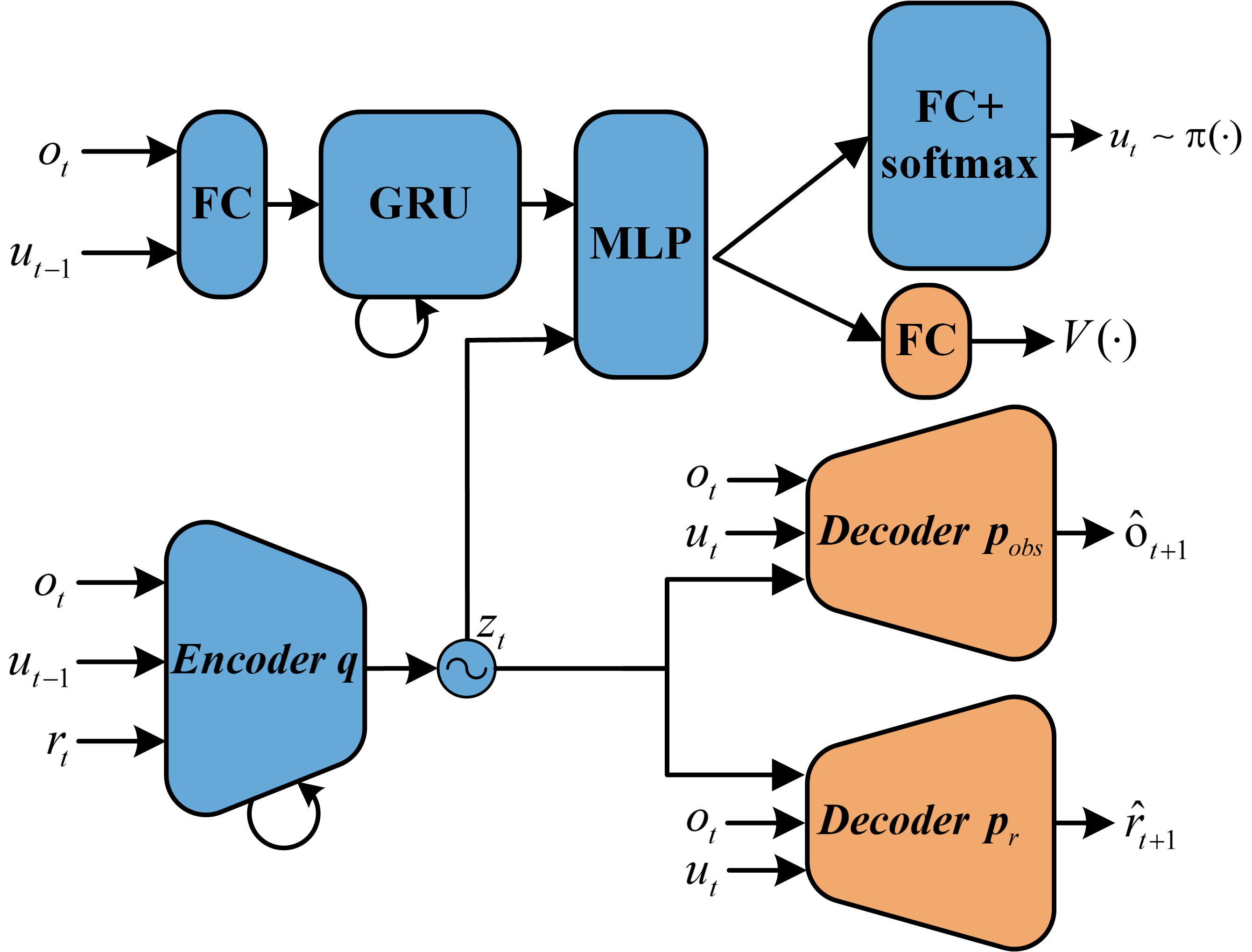}
	\caption{The architecture of Fact-based Agent Modeling (FAM). During the execution phase, the encoder module of FBI utilizes the local information of the agent to extract representations of other agents' policies, which are then used for the agent's decision-making process. On the other hand, during the training phase, the decoder module reconstructs the facts and simultaneously trains both the encoder and decoder. The agent makes decisions based on its own action-observation trajectories and the representations of other agents' policies.}
	\label{fig:fam_architecture}
\end{figure}

In this section, we introduce a Fact-based Agent Modeling (FAM) Algorithm as shown in Figure \ref{fig:fam_architecture}, which completely eliminates the assumption that traditional agent modeling allow access to other agents local information during the training or execution phases. Firstly, we provide the structure and details of fact-based belief inference module (FBI). Furthermore, we present the optimization objective and training procedure for FAM.


\subsection{Fact-based Belief Inference}

To enable an agent to interact with other agents and learn adaptive policy, it needs to infer the current policies of the interacting agents. Fact-based Belief Inference (FBI) eliminates the assumption that agents can access other agents' local information during training or execution. This module extracts policy representation denoted as $z^{i}$ of other agent from the interaction trajectories of agent $i$, including triplets of observations, actions, and rewards triplets. Policy representations are learned from facts acquired by the agent after performing actions. The extracted representation $z^{i}$ denotes agent $i$'s beliefs about other agents, i.e., estimates of their policy. This introduces uncertainty of other agents' policies into agent $i$'s policy $\pi_{i} (a_{t}^{i}|\tau_{1:t}^{i},z^{i})$.

Assuming the joint policy of other agents are unobservable variables $z^{i}$ in the latent space $\mathcal{Z}^{i}$ for agent $i$, and the latent variable $z_{t}^{i}$ at time step $t$ contains the policy representation of other agents except agent $i$ itself. To learn the latent information, FAM employs FBI which is a variational encoder-decoder architecture\cite{kingma2013auto} as shown in Fig.\ref{fig:fam_architecture}. Agent $i$ uses an encoder $q$ consists of a recurrent neural network and a fully connected neural network to infer representations of other agents' policies by local information  including the observation-action-reward triplet. It outputs $\mu^{i}$ and $\log(\sigma^{i})$ which is the parameters of variational distribution. And the policy representation of other agents are sampled from the variational distribution. 
Specifically, the goal of the encoder is to approximate the true posterior $p(z^{i})$ using a variational distribution obtained solely from local information. FBI constructs a decoder $p$ to learn the representation of other agents by reconstructing facts conditioned on policy representations $z^{i}\sim \mathcal{N}(\mu^{i},\sigma^{i})$ and the agent's observation-action. The encoder is parameterized by $\psi^{i}$, and the state prediction and reward prediction function in the decoder are parameterized by $\phi^{i}$ and $\varphi^{i}$, respectively. FBI jointly optimizes $\psi^{i}$, $\phi^{i}$, and $\varphi^{i}$ to maximize the evidence lower bound(ELBO) of the sampled trajectory $\tau_{1:t}^{i}$, as follows,

\begin{equation}
	\begin{aligned}
		\label{eq:elbo}
		ELBO(\psi^{i}, \phi^{i}, \varphi^{i} | \tau_{1:t}^{i}) 
		= \mathbb{E}_{z_{t}^{i}\sim q_{\psi^{i})}} \big[
		J_{recon}
		\big] \\
		- \beta D_{KL}(q(\mu_{t}^{i},\sigma_{t}^{i} |\tau_{1:t}^{i};\psi^{i}) || p(z^{i}))
	\end{aligned}
\end{equation}
where $J_{recon}=\log p(\hat{o}_{t+1}^{i},\hat{r}_{t+1}^{i}|o_{t}^{i},a_{t}^{i},z_{t}^{i};\phi^{i},\varphi^{i})$ is reconstruction loss. The ELBO is related to the state transition and reward functions of each agent $i$. $\tau_{1:t}^{i}=(o_{1}^{i},a_{1}^{i},r_{2}^{i},o_{2}^{i},...,r_{t}^{i},o_{t}^{i})$ represents the local trajectory information of agent $i$ up to time step $t$. $p(z^{i})$ is the prior distribution of the latent variable $z^{i}$, we assume the latent variable follows a standard Gaussian distribution $z^{i}\sim\mathcal{N}(0,I)$. $D_{KL}$ is the Kullback-Leibler(KL) divergence which measures the distance between the approximate posterior distribution $q$ and the true posterior distribution $p$. The hyperparameter $\beta$ is used for controling the importance of the regularization term $KL$ divergence\cite{higgins2017beta}. 
Minimizing the loss is equivalent to maximizing the ELBO, and thus the loss function of the FBI as follows:

\begin{equation}
	\begin{aligned}
		\label{eq:fam_fbi_loss}
		\mathcal{L}_{fbi}(\psi_{i},\phi_{i},\varphi_{i})= \mathbb{E}_{z_{t}^{i}\sim q_{\psi^{i}}} \big[ 
		J_{recon\_obs} + J_{recon\_rew}\big] \\
		- \beta \frac{1}{2} \sum_{j=1}^{d}(1+\log(\sigma_{t,i,j}^{2})-\mu_{t,i,j}^{2}-\sigma_{t,i,j}^{2})
	\end{aligned}
\end{equation}
where $J_{recon\_ obs}=(p_{obs}(\hat{o}_{t+1}^{i}|o_{t}^{i},a_{t}^{i},z_{t}^{i};\phi^{i}) - o_{t+1}^{i})^{2}$ and $J_{recon\_ rew}=(p_{r}(\hat{r}_{t+1}^{i}|o_{t}^{i},a_{t}^{i},z_{t}^{i};\varphi^{i}) - r_{t+1}^{i})^{2}$ are the observation prediction and reward prediction loss functions, respectively.
$d$ represents the dimensionality of the latent variable $z^{i}$. The intuitive interpretation of this loss function is that the decoder $p$ is optimized to reconstruct the facts that occur after taking an action, specifically received in next time step's observation and reward.

\renewcommand{\algorithmicrequire}{\textbf{Initialize:}}  
\renewcommand{\algorithmicensure}{\textbf{Output:}} 
\begin{algorithm}[!t]
	\caption{Training Procedure for FAM Algorithm} 
	\label{alg:fam}
	\begin{algorithmic}[1]
		\Require
		$\theta_{i},\omega_{i},\Phi_{i}=\{\psi_{i},\phi_{i},\varphi_{i}\},\beta,\alpha_{1},\alpha_{2},\mathcal{B}, E$;
		\Ensure
		$\theta_{i}^{*},\omega_{i}^{*},\Phi_{i}^{*}=\{ \psi_{i}^{*},\phi_{i}^{*},\varphi_{i}^{*} \}$;
		\For {each episode $j$}
		\State Initial observation $\mathbf{o}_{0} \leftarrow\{o_{0}^{i}\}_{1}^{n}$
		\For {each timestep $t$}
		\State Get observations $\textbf{o}_{t}=\{o_{t}^{i}\}_{i=1}^{n}$;
		\State Compute opponent embeddings $\textbf{z}_{t}=\{z_{t}^{i}\sim q_{\psi_{i}}\}_{i=1}^{n}$;
		\State Sample action $u_{t}^{i}\sim\pi_{i}(\cdot|o_{t}^{i},z_{t}^{i};\theta_{i})$;
		\State Perform joint actions $\textbf{u}_{t}=(u_t^1,...,u_t^n)$ and reveive joint reward $r_{t+1}$ and next observations
		\State $\textbf{o}_{t+1}=\{o_{t+1}^{i}\}_{i=1}^{n}$;
		\State Add transition $\{ \textbf{o}_{t}, \textbf{z}_{t}, \textbf{u}_{t},r_{t+1},\textbf{o}_{t+1}\} \rightarrow \mathcal{B}_{j,t}$
		\EndFor
		\If {$|\mathcal{B}|$ = batch size}
		\For {each epoch $e < E$}
		\For {each agent $i$}
		\State $\omega_{i} \leftarrow \omega_{i} - \alpha_{1} \nabla_{\omega_{i}} \mathcal{L}_{critic}$ (Eq.\ref{eq:fam_critic_loss})
		\State $\theta_{i} \leftarrow \theta_{i} - \alpha_{1} \nabla_{\theta_{i}} \mathcal{L}_{actor}$ (Eq.\ref{eq:fam_actor_loss})
		\State $\Phi_{i} \leftarrow \Phi_{i} - \alpha_{2} \nabla_{\Phi_{i}} \mathcal{L}_{fbi}$ (Eq. \ref{eq:fam_fbi_loss})
		\State $B \leftarrow \emptyset$;
		\EndFor
		\EndFor
		\State Soft update parameters $\theta_{i}^{'},\omega_{i}^{'},\Phi_{i}^{'}$ with $\theta_{i},\omega_{i},\Phi_{i}$
		\EndIf
		\EndFor
	\end{algorithmic}
\end{algorithm}
\subsection{Training Algorithm of FAM}
In this section, we describe the training process of FAM. The sampled trajectory of the agent, along with the latent variable $z^{i}$, is used to optimize the RL policy. We consider an augmented policy space $\mathcal{T}{aug}^{i}=\mathcal{O}^{i}\times \mathcal{U}^{i}\times\mathcal{Z}^{i}$ for agent $i$, where $\mathcal{O}^{i}$ and $\mathcal{U}^{i}$ are the original observation and action spaces of the Dec-POMDP, and $\mathcal{Z}^{i}$ represents the belief space of agent $i$ on other agents. Specifically, the belief refers to the policy representations of other agnets. Compared to the policy space without considering other agents' policy representations $\mathcal{T}{aug}^{i}=\mathcal{O}^{i}\times \mathcal{U}^{i}$, the augmented policy space $\mathcal{T}^{i}$ allows for different responses to different $z^{i}\in\mathcal{Z}^{i}$. This enables adaptive behavior based on the policy representations of other agents. We assume that all agents learn simultaneously in the same environment. Due to the delayed nature of other agents' policy changes, which affect the agent's belief about their policy representations, we train the FAM using on-policy algorithm. In our experiments, we use the PPO algorithm to optimize the agent's policy. The inputs to the Actor and Critic are the local action-observation trajectories and the inferenced policy representation $z^{i}$. Additionally, the RL loss does not backpropagate into FBI. To encourage exploration, we also use policy entropy\cite{mnih2016asynchronous}. Given a batch of sampled trajectories $\mathcal{B}$, the loss for the Critic network is defined as follows:

\begin{equation}
	\begin{aligned}
	\label{eq:fam_critic_loss}
	\mathcal{L}_{critic}(\omega_{i})
	= \mathbb{E}_{\mathcal{B}}\big[
	(r_{t+1}^{i}+\gamma \overline{V}(o_{t+1}^{i},\overline{z}_{t+1}^{i};\omega_{i}^{-})\\
	-V(o_{t}^{i},\overline{z}_{t}^{i};\omega_{i}))
	\big]
	\end{aligned}
\end{equation}
where $\overline{V}$ is the target network, $\overline{z}$ indicates that the loss of the Critic network does not backpropagate through $z$, and $\omega^{-}$ represents the parameters of the Critic target network, which are also not updated through gradient backpropagation. Additionally, the loss for the Actor network is defined as follows:

\begin{equation}
	\begin{aligned}
	\label{eq:fam_actor_loss}
		\mathcal{L}_{actor}(\theta_{i})
		= \mathbb{E}_{(o_{t}^{i},a_{t}^{i},z_{t}^{i},r_{t+1}^{i},o_{t+1}^{i})\sim \mathcal{B}}\big[ \\
		\min(\text{ratio}_{t}(\theta_{i}), \text{clip}(\text{ratio}_{t}(\theta_{i}), 1-\epsilon, 1+\epsilon))A_{t}^{i}
		\big]
	\end{aligned}
\end{equation}
where $\text{ratio}_{t}=\frac{\pi(a_{t}|o_{t},z_{t};\theta^{i})}{\pi(a_{t}| o_{t},z_{t};\theta_{old}^{i})}$ is the action probability ratio of the old and new policies, and $\text{clip}$ modifies the surrogate objective by restricting the probability ratio.

Now we can define a training objective to learn the approximate posterior distribution $q$ as well as reward prediction function $p_r$ and observation prediction function $p_{obs}$, as follows:
\begin{equation}
	\begin{aligned}
		\label{eq:fam_loss}
		\mathcal{L}_{fam}(\theta,\omega,\psi,\phi,\varphi)\!\! \!=\ \ \mathcal{L}_{actor} + \mathcal{L}_{critic} + \mathcal{L}_{fbi}
	\end{aligned}
\end{equation}

\section{Experimental Results and Analysis}
In this section, we aim to investigate the following aspects of FAM: 1). whether FBI improve learning efficiency, promote cooperative among multi agents, and learn adaptive collaboration strategies? 2). how FAM enables collaboration through adaptive strategies, 3). does FBI encode the policies of other agents, and beyong that, what other important information is encoded. To answer these three questions, we conduct experments in two multi-agent particle environments: Cooperative Navigation (CN) and Predator-Prey (PP), as introduced by Lowe et al.\cite{lowe2017multi}. The implementation is based on the epymarl framework proposed by Papoudakis et al.\cite{papoudakis2020benchmarking}.

\subsection{Experimental Setup}

\subsubsection{Environments}
We introduce the two environments used for our proposed FAM and baselines.

\textbf{Cooperative Navigation}: In this task, $N$ agents need to cooperatively occupy $L$ landmarks in an environment with partial observability. The team reward function $r_{distance} = -\sum_{i=1}^{N}\min_{j} dis(Landmark_{i},Agent_{j})$ is based on the negative sum of distances between the landmarks and their closest agents. Additionally, we discourage collisions between agents, and a collision penalty $r_{collision} = -1$ is applied. Each agent needs to observe the movements of other agents to infer their behavior or goals, select a suitable landmark to occupy while avoiding conflicts and collisions with other agents. We conduct  experiments with $N=5$ agents and $L=5$ landmarks, where each agent can only observe the relative positions of the two closest agents and three closest landmarks. The agents have $5$ available actions, and the maximum episode length is set to $25$.

\textbf{Predator-Prey (PP)}: In this task, $N$ predators try to capture $M$ preys in an environment with partial observability. The preys follow predefined policies to move away from the closest predators with a speed of $7$, while the predators are only $5$. Since preys have a faster movement speed than predators, individual predators cannot capture preys on their own. The team reward for predators is the negative sum of distances between the preys and their closest predators. A collision penalty $r_{collision} = -1$ is also applied to the predators. In PP task, each predator can observe the relative positions of the two closest predators and three closest preys. The predators have $5$ available actions. Unlike in Cooperative Navigation, the preys exhibit highly dynamic movement. Therefore, predators need to infer the behavior of other predators and target preys to cooperate with other predators and capture the desired preys.  We conduct  experiments with $N=7$ predators and $M=3$ preys.


In IA2C, each agent treats other agents as a part of the environment and utilizes the A2C algorithm to learn and optimize its Actor network with \ref{eq:a2c_actor_loss} for approximating the policy $\pi_{\theta_{i}}$ and Critic network with \ref{eq:a2c_critic_loss} for approximating the value function $V_{\omega_{i}}$ in distributed multi-agent systems\cite{mnih2016asynchronous}.

\subsubsection{Implementation Detials}
Next, we will introduce the implementation details of the proposed FBI and FAM. The FAM consists of actor network, critic network and FBI network parameterized by $\theta_i$, $\omega_i$ and $\Phi_i = \{\psi_i, \phi_i, \varphi_i\}$, respectively. The FBI includes an RNN-based encoder and two MLP-based decoders. During execution, RNN-based encoder takes the local observation-action-reward triplet ($o_t^i, u_{t-1}^i, r_t^i$) as input through a 1-layer fully connected neural network(FC) followed by a ReLU activation function to extract features, which are then fed into a GRU recurrent network to capture temporal dependencies. Finally, a 1-layer FC outputs the variational distribution parameters $\mu_t^i$ and $\log \sigma_t^i$ that approximate the true posterior distribution. The sampled with dimension $d=5$ is the policy representation of other agents for agent decision-making. During training, MLP-based decoder takes the local observation-action-policy representation ($o_t^i, u_t^i, z_t^i$) as input through 3-layer FC and followed by ReLU activation functions to output the predictions of rewards and observations obtained after executing action. It is important to note that the last fully connected layer does not require a ReLU activation function. The RNN-based encoder and MLP-based decoder are trained by computing the prediction loss and regularization term. 


\begin{figure*}[!t]
	\centering
	\includegraphics[width=0.9\linewidth]{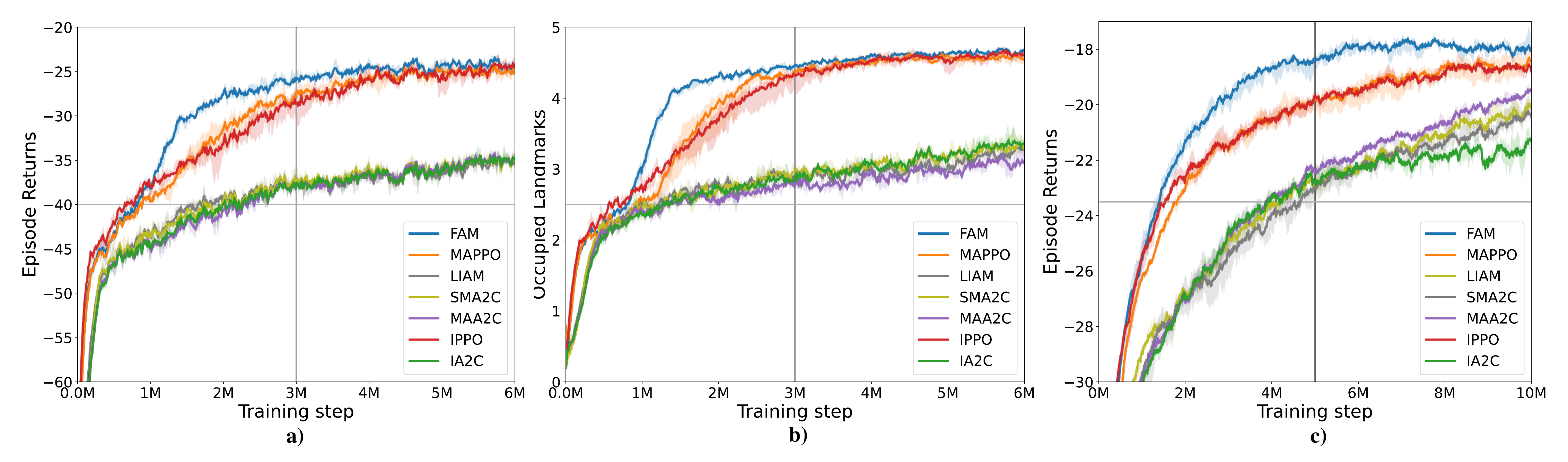}
	\caption{The training results of the proposed FAM and the baseline method in CN and PP environments. a)The episode return curves during training in CN. b)The occupied landmarks curves during training in CN. c)The episode return curves in PP. The solid line is the mean of the  training results of 5 random seeds, and the shaded area is the 25\%-75\% quartile.}
	\label{fig:main_results}
\end{figure*}

\begin{table}[!t]
	\centering
	\caption{Performance evaluation comparison of FAM against baselines in CN.}
	\label{tab:algorithm}
	\begin{tabular}{c cccc}
		\hline\noalign{\smallskip}
		Methods & Avg. Ret. & Avg. Rew. & Avg. Occ. & Avg. Dist.\\
		\noalign{\smallskip}\hline\noalign{\smallskip}
		IPPO 	& $-26.2\pm 8.2$  & $-0.52\pm 0.3$ &  $4.56\pm 0.4$ & $\textbf{0.45}\pm 0.2$ \\
		IA2C 	& $-34.6 \pm 10.1$ & $-1.08\pm 0.5$ &  $3.48\pm 1.0$ & $1.02\pm 0.4$ \\
		MAA2C 	& $-34.8\pm 9.4$ & $-1.09\pm 0.4$ & $3.06\pm 1.1$ & $1.06 \pm 0.4$ \\
		MAPPO 	&  $-26.1\pm 7.8$ & $-0.52\pm 0.3$ &  $4.57\pm 0.5$ & $0.48\pm 0.3$ \\
		FAM & $-\textbf{25.5}\pm 8.8$ & $-\textbf{0.50}\pm 0.3$ & $\textbf{4.45}\pm 0.5$ & $\textbf{0.45}\pm 0.3$ \\
		LIAM 	& $-35.2\pm 9.6$ & $-1.06\pm 0.4$ & $3.31\pm 1.0$ & $1.05\pm 0.4$ \\
		SMA2C 	& $-35.3\pm 9.6$ & $-1.12\pm 0.5$ & $3.26\pm 1.1$ & $1.05\pm 0.4 $\\
		\noalign{\smallskip}\hline
	\end{tabular}
	\label{tab:cn_performance}
\end{table}

\subsection{Main Results}
We compare FAM with several baselines to verify the effectiveness and feasibility of the proposed method. Figure \ref{fig:main_results}a) and Figure \ref{fig:main_results}b) show the average episode return curves and average landmarks occupied curve of FAM compared with other baseline algorithms in CN during training. Figure \ref{fig:main_results}c) shows the average episode return curves of various methods trained in PP with training duration of 1e7 steps. And Table \ref{tab:cn_performance} presents the performance metrics of various algorithms evaluating 100 episodes in CN, including average episode return(Avg. Ret.), average reward at the final timestep(Avg. Rew.), average occupied landmarks(Avg. Occ.), and the sum of the average distances of all landmarks from the nearest agent(Avg. Dist.).

From Fig. \ref{fig:main_results} a). and \ref{fig:main_results} b)., we can see that the proposed FAM achieves higher learning efficiency than all other baselines from $1e6$ training steps, as well as faster convergence, and slightly outperforms IPPO and MAPPO from $4e6$ training steps. The main reason is that after all agents learn a certain strategy, considering the strategies of other agents will help the agents learn adaptive cooperation strategies. And they struggle with partial observability of the environment after 4e6 steps. Meanwhile, the shaded areas of IPPO and MAPPO are larger compared with FAM, which indicates that the cooperative strategy without considering other agents is less robust. Among the four evaluation performances as shown in Table \ref{tab:cn_performance}, FAM has reached the best compared with all other baselines. The average episode return curves of IA2C, MAA2C, LIAM, and SMA2C overlap and show slow learning. And in Figure \ref{fig:main_results}b), IA2C occupies more landmarks. Similar results are shown in Table \ref{tab:cn_performance} which indicates that IA2C performs slightly better than SMA2C, LIAM, and MAA2C. The possible reason is that the low sample efficiency of the A2C methods and agents only need to consider other agents at certain critical moments in CN, which makes the performance of independent similar to CTDE methods. Additionally, LIAM and SMA2C do not show superiority. This could be due to the low sample efficiency of the A2C methods and the high randomness in directly modeling the actions of other agents in a non-stationary environment.

It can be seen from Figure \ref{fig:main_results}c) that IA2C performs the worst and the proposed FAM outperforms all other baseline by considering the strategies of other agents. The possible reason is that preys are highly dynamic and move faster than predators, which requires closer cooperation between predators and adaptation to other predators' strategy to capture preys cooperatively. However, independent IA2C is difficult to achieve. Moreover, MAA2C employs a centralized critic that utilizes global information to guide the policy learning of agents and achieves better performance than IA2C. But the performance of LIAM and SMA2C falls behind the centralized critic. The possible reason is that the centralized critic provides more effective information for guiding policy learning under partially observability. IPPO and MAPPO exhibit similar average episode return curves which can be attributed to the effectiveness of the PPO algorithm. This finding aligns with previous studies\cite{yu2022surprising} and \cite{papoudakis2020benchmarking}.

In general, in the experimental settings of CN andPP, we found it interesting that the agent modeling method can quickly and effectively improve the learning efficiency and learn adaptive collaboration strategies to obtain higher rewards after other agents learn a certain strategy. The good news is that it didn't hinder the agent's strategy learning before this.

\begin{figure}[!t]
	\centering
	\includegraphics[width=0.8\linewidth]{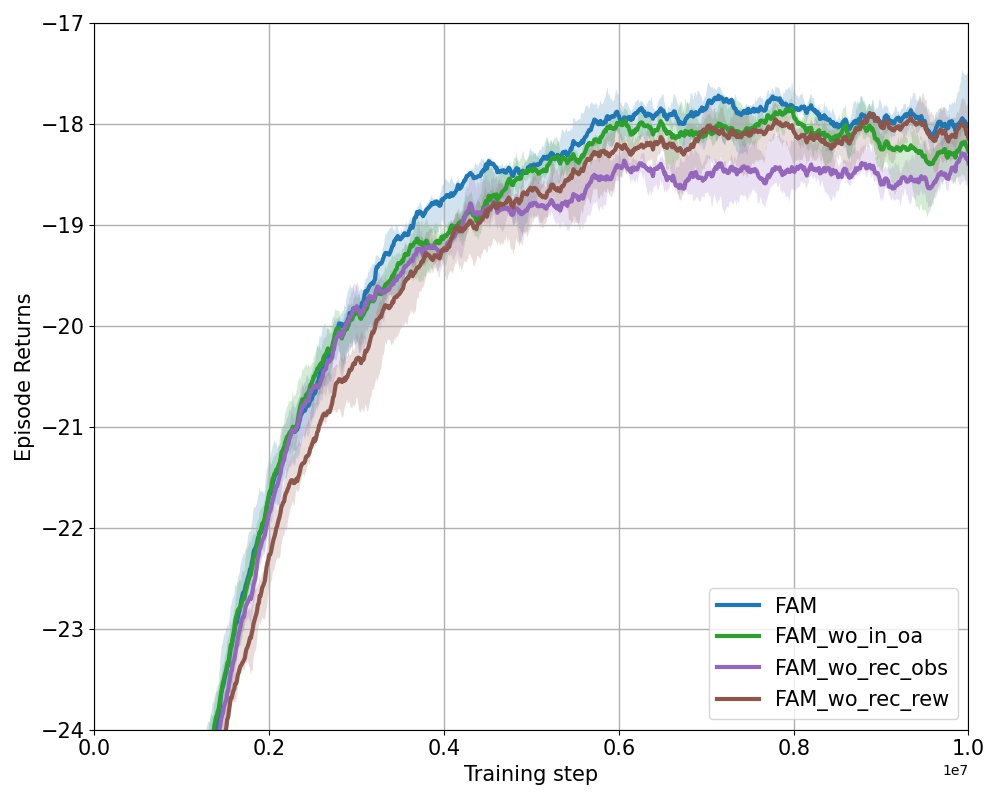}
	\caption{Comparing the ablation results of FAM and its belief inference network in the Predator-Prey environment.}
	\label{fig_pdf:ablation_study}
\end{figure}
\subsection{Ablation Results}
The fact-based belief inference (FBI) network in FAM utilizes a variational autoencoder (VAE) architecture, whose input and reconstruction target are the local information of the agent. To investigate the impact of the following factors in FBI network: 1) decoder input and 2) reconstruction targets, we conducted ablation experiments in the Predator-Prey environment.

The ablation experiments included the following variations: i). FAM\_wo\_in\_oa, where the decoder input only consisted of the representation of other agents' policies. ii). FAM\_wo\_rec\_obs, where the decoder only reconstructed rewards. iii). FAM\_wo\_rec\_rew, where the decoder only reconstructed observations. The average episode return curves of the Predator-Prey task are plotted in Figure \ref{fig_pdf:ablation_study}. These ablation experiments aim to examine the contributions of different components in FBI. By comparing the performance of these variations with the FAM, we can gain insights into the importance of decoder input and the reconstruction of observations and rewards.

The decoder in FBI network takes the agent's local observations, actions and inferred representation as input. We denote the agent's local observations and actions as "oa". To compare the impact of the decoder input, we denote the decoder input as FAM\_wo\_in\_oa only for other agent policy representations $z$ and keep the reconstruction fact unchanged. As shown in Figure \ref{fig_pdf:ablation_study}, FAM is generally better than FAM\_wo\_in\_oa. Although FAM and FAM\_wo\_in\_oa have similar performance in the early stage, both can effectively improve the efficiency of policy learning. But when struggling partial observability, FAM has an advantage. The possible reason is that the decoder design of FBI is better for the agent to understand the dynamics of the environment.

To compare the impact of the decoder reconstruction targets, we compare the training results of FAM\_wo\_rec\_obs and FAM\_wo\_rec\_rew on PP task, as shown in Figure \ref{fig_pdf:ablation_study}. It can be seen that the performance of FAM\_wo\_rec\_rew is better, but it is weaker than FAM\_wo\_rec\_obs in the early stage. The possible reason is that the team reward helps to extract other agents' policy representations, but this is a spurious reward signal, which may also hinder the learning of other agents' policies. The observation can directly represent the movement information of the surrounding agents, which provides rich verification information for each individual. Moreover, FAM has the advantages of FAM\_wo\_rec\_rew and FAM\_wo\_rec\_obs by reconstructing observations and rewards, and has the best performance both in the early stage of training and in the stage of struggling partial observability.

Overall, the decoder reconstructs observations and rewards by inputting its own observations, actions, and policy representations to better help the agent understand the dynamics of the environment.

\begin{figure*}[!t]
	\centering
	\includegraphics[width=0.9\linewidth]{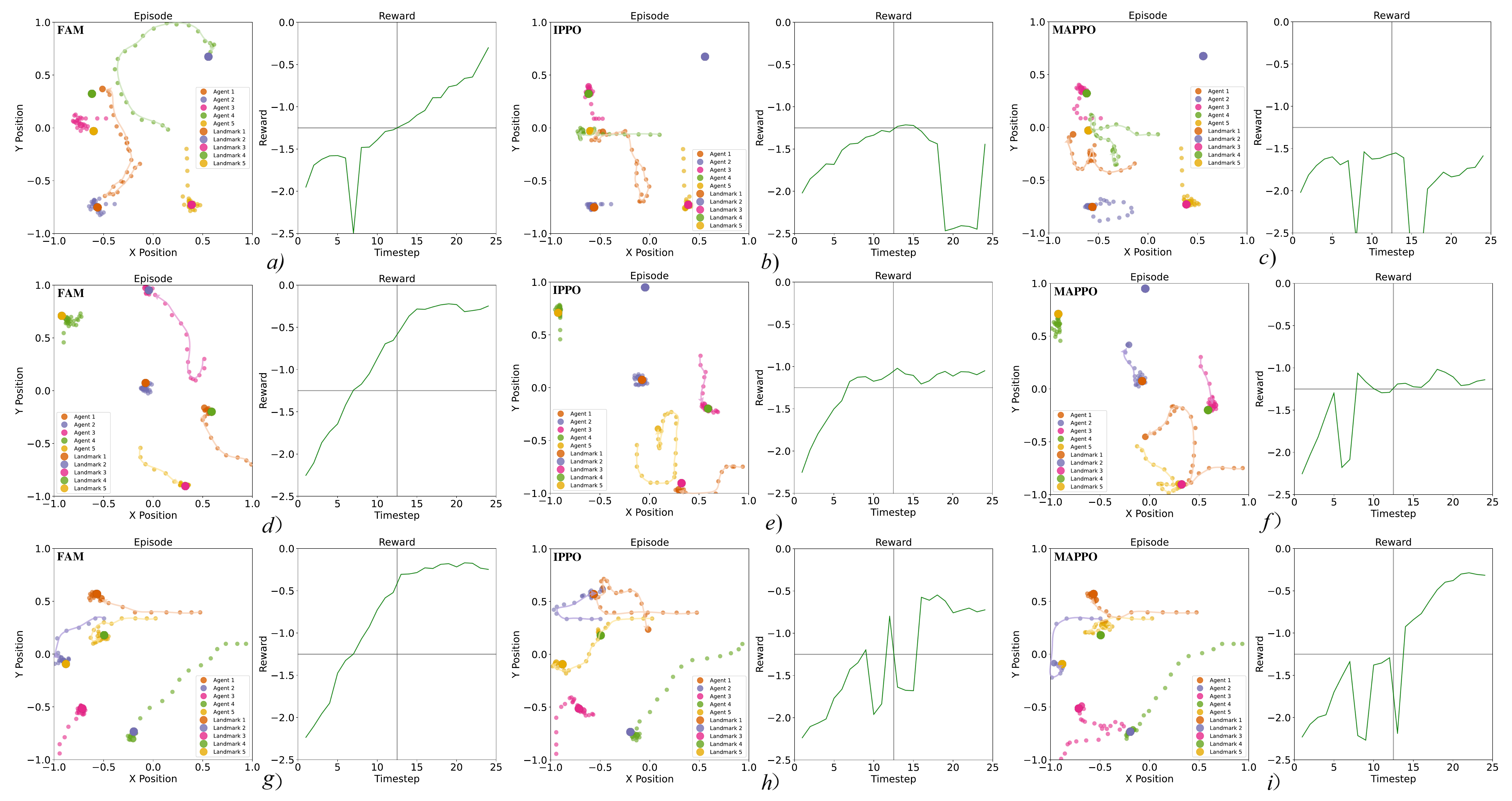}
	\caption{The navigation trajectories and immediate reward curves of three evaluation scenarios of FAM, IPPO and MAPPO under Cooperative Navigation task.}
	\label{fig_pdf:strategy_analysis}
\end{figure*}
\subsection{Strategy Analysis}

In order to understand the representation of other agents learned by FAM, we conducted evaluations and compared and analyzed the cooperative strategies of FAM agents with IPPO and MAPPO agents in the Cooperative Navigation. Figure \ref{fig_pdf:strategy_analysis} shows the evaluated navigation trajectories and immediate reward curves of the three methods of FAM, IPPO and MAPPO in CN.

It can be seen that all FAM agents can move near the landmark or successfully occupy the landmark, and the immediate rewards are the best. However, the IPPO agent and the MAPPO agent can not due to the goal conflict between the agents. We summarize the cooperative skills learned by FAM agents, including i). Communicate without communication (CWC), ii). Avoiding goal conflict and competition (AGCC), iii). Giving up the small to keep the big (GSKB).

\textbf{Communicate without communication}: 
Once the strategy of other agents is found to change, it will change its own strategy in time to meet the needs of the task. As shown in Figure \ref{fig_pdf:strategy_analysis}a), both agent 1 and agent 2 want to occupy landmark 1, and there is a goal conflict. However, agent 2 has an advantage in distance when occupying landmark 1, so agent 1 has to change the landmark to 2. At this time, Agent 4 wants to occupy landmark 4, but it infers that the strategy of Agent 1 is occupy landmark 2 which is farther away, which makes it take longer to complete the task. There, agent 4 changes the navigation landmark to 2 and agent 1 changes its own navigation landmark to 4 for shorest complete time. However, Agent 1 and Agent 4 of IPPO and MAPPO have landmark conflict and competition, as shown in Figure \ref{fig_pdf:strategy_analysis} b) and c).

\textbf{Avoiding goal conflict and competition}: 
When it is found that the goals of other agents conflict with itself, it will change its own goals according to the actual situation to avoid competition. As shown in Figure \ref{fig_pdf:strategy_analysis}d), there is a goal conflict between agent 1 and agent 3 bacause they want to occupy landmark. However, agent 1 occupys landmark 4 is more advantageous, because agent 3 is closer to unoccupied landmark 2. Therefore, agent 3 changes the landmark to avoid goal conflict can promote overall cooperation and complete the task faster. In contrast, IPPO agent and MAPPO agent failed to achieve this. As shown in Figure \ref{fig_pdf:strategy_analysis}e), there is a goal conflict and goal competition between Agent 1 and Agent 5. Agent 5 cannot observe landmark 2 due to partial observability. Therefore, agent 5 cannot effectively occupy landmark. A possible effective method is that agent 2 changes its own landmark to 2. This collaborative strategy is reflected by  MAPPO, as shown in Figure \ref{fig_pdf:strategy_analysis}f). But it takes longer to complete the task.

\textbf{Give up the small to keep the big}: When there is a goal conflict or the goal can be occupied by a more advantageous agent, the agent will change its own goal to shorten the time to complete the task. This skill is demonstrated in Figure \ref{fig_pdf:strategy_analysis}g), where Agent 2, although closer to landmark 1 and 4, chooses the farther landmark 5 to facilitate faster landmark occupation by Agent 1 and Agent 5. In contrast, IPPO and MAPPO agents fail to achieve the goal of conflict-free. In Figure \ref{fig_pdf:strategy_analysis}h) and \ref{fig_pdf:strategy_analysis}i), there is a landmark conflict and competition between agents.

\begin{figure*}[!t]
	\centering
	\includegraphics[width=0.9\linewidth]{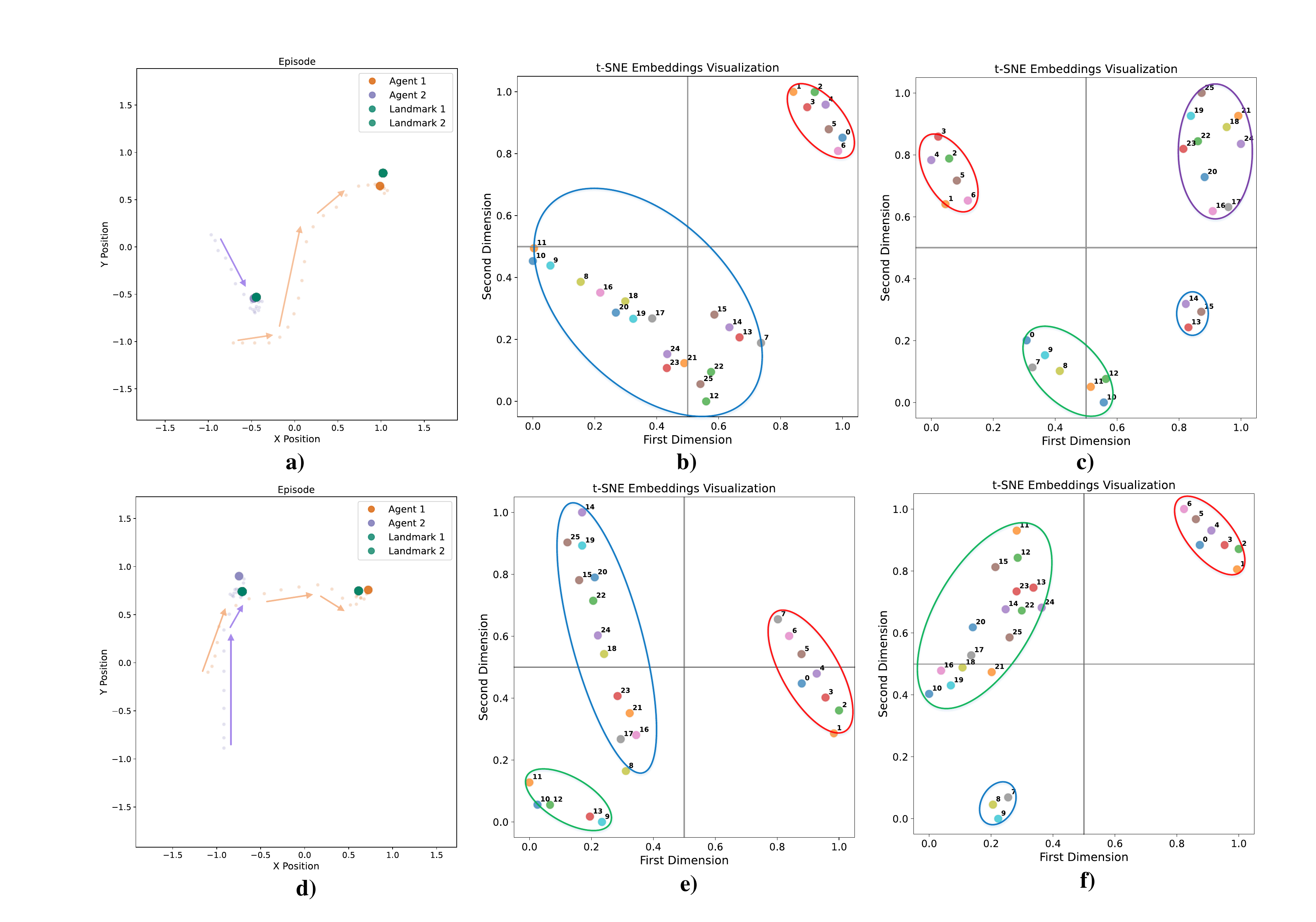}
	\caption{a) and d) are trajectory plots of the agents in the CN. b) and c) are t-SNE projections of the embedding vectors learned by agent 1 and agent 2 in trajectory a). And e) and f) are t-SNE projections of the embedding vectors learned by agent 1 and agent 2 in trajectory b), respectively. We use a maximum time step of 25 to visualize the embedding vector of each time step, and use circles with different colors for each cluster.}
	\label{fig_pdf:encoder_evaluation}
\end{figure*}

\subsection{Encoder Evaluation}
After evaluating the advantages of the FBI module in FAM for adaptive strategy learning, we analyzed the embedding vectors learned by the RNN-based encoder in FBI to gain deeper insights into the proposed method. We addressed the question of whether FAM encode the strategies about other agents. We visualized the embedding vectors of the RNN-encoder and analyzed the learned embeddings. To facilitate the understanding of the encoded embeddings of other agents, we conducted experiments in CN with $N=2$ agents, $L=2$ landmarks. Figure \ref{fig_pdf:encoder_evaluation} visualizes the evaluation results.

We observed that points corresponding to adjacent time steps tend to form clusters, and each cluster is correlated with the agent's motion state. From Figure \ref{fig_pdf:encoder_evaluation}a), we can see that Agent 2 moves towards the bottom right direction, approaching the landmark and hovering around it. These two processes form two distinct clusters in Figure \ref{fig_pdf:encoder_evaluation}b). And Agent 1's motion consists of four steps, with the first three steps marked by arrows and the final step involving hovering around the landmark. These four steps correspond to the four distinct clusters formed in Figure \ref{fig_pdf:encoder_evaluation}c). Based on these observations, we hypothesize that different clusters represent different aspects of the modeled agent's motion, including the magnitude and direction of motion.

Additionally, we speculate that the encoder embedding vectors also include the positional information of the modeled agents. From the trajectory in Fig .\ref{fig_pdf:encoder_evaluation}d), it can be seen that the agent 2 moves upward first, then moves upward to the right and gradually approaches landmark and hovers around it. These three processes also form three different clusters in Fig .\ref{fig_pdf:encoder_evaluation}e). Compared to the red cluster, the blue cluster is closer to the green cluster. We can see that there are only three clustering results in Fig.\ref{fig_pdf:encoder_evaluation}f), and its motion process has four steps. The possible reason is that the last movement close to the landmark is close to the position hovering near the landmark, and they are classified into the same cluster. In addition, in the same cluster, the distance between points at adjacent moments is small, while the distance between points at multiple moments is large.

\section{Conclusion}
We have proposed a Fact-based Agent Modeling (FAM) for multi-agent learning that build FBI to reconstruct facts for achieving agent modeling without accessing local information of other agents. By considering the policy of other agents during decision-making, FAM outperforms baseline methods and achieving higher rewards in complex mixed scenarios. Extensive experimental is conducted to verify the effectiveness and feasibility of the proposed FAM and analyse the encoder information of FBI.

\bibliographystyle{IEEEtran}
\bibliography{Literature.bib}
\begin{IEEEbiography}[{\includegraphics[width=1in,height=1.25in,clip,keepaspectratio]{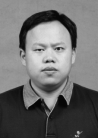}}]{Baofu Fang}
	received Ph.D. degree in Computer Application Technology from Harbin Institute of Technology, China in 2013. He joined Department of Computer Science and Technology, School of Computer Science and Information Engineering, Hefei University of Technology in 2000, and An Associate Professor in 2010, and Master’s Supervisor in 2011. His current research interests include multi robot/agent system, emotion/self-interest robot and machine learning. He is the Technology Chair of Anhui Robot Competition, Member of Standing Committee of China Association of Artificial Intelligence (CAAI) Young Committee, Member of Standing Committee of China Association of Artificial Intelligence (CAAI) Robot and Culture Committee.\end{IEEEbiography}

\begin{IEEEbiography}[{\includegraphics[width=1in,height=1.25in,clip,keepaspectratio]{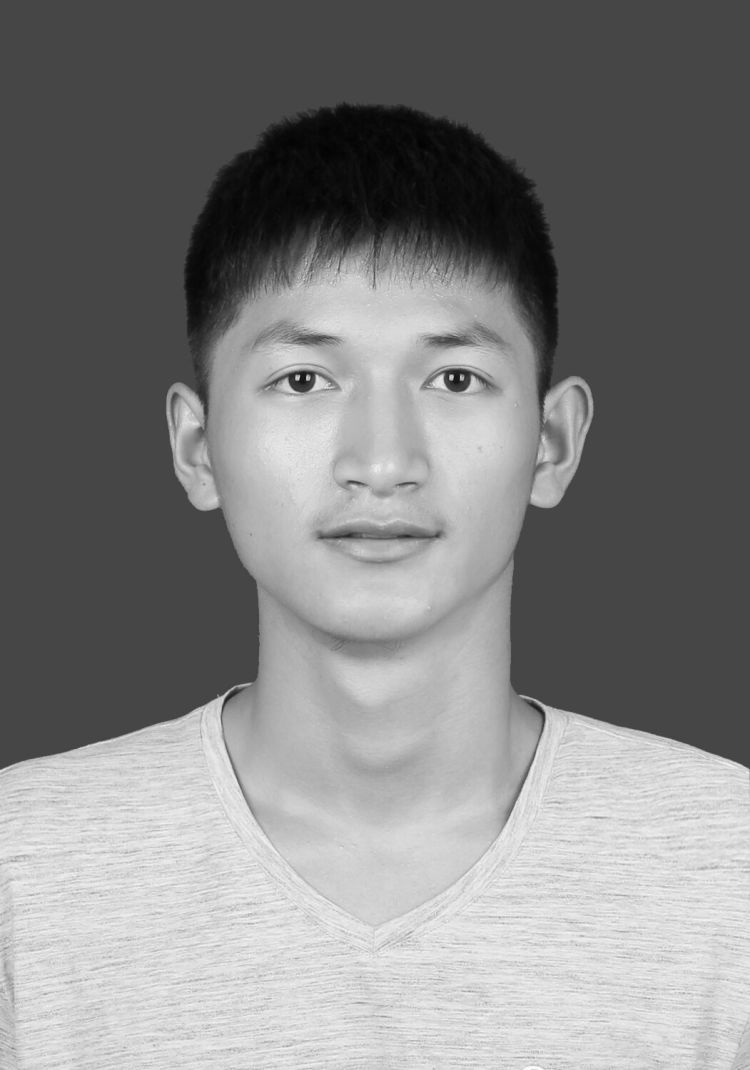}}]{Caiming Zheng}
	received the B.Eng degree in computer science and technology from Ningbo University of Technology, China in 2021. He is currently pursuing a M.S. degree in computer science and technology at the School of Computer Science and Information Engineering, Hefei University of Technology, China. His research interests include multi-agnet systems, reinforcement learning and multi-agent reinforcement learning.\end{IEEEbiography}

\begin{IEEEbiography}[{\includegraphics[width=1in,height=1.25in,clip,keepaspectratio]{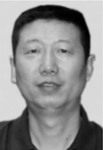}}]{Hao Wang}
	received the B.Eng degree from Shanghai Jiao Tong University in 1984, and received M.S. degree and Ph.D. degree from Hefei University of Technology in 1989 and 1997, respectively. He is currently a Professor and Doctoral Supervisor with the School of Computer Science and Information Engineering, Hefei University of Technology. His research interests include intelligent computing theory and software, distributed intelligent systems, complex system theory and modeling, etc.\end{IEEEbiography}
	
\end{document}